\RequirePackage{fix-cm}
\documentclass[twocolumn]{svjour3}          
\smartqed  
\usepackage{graphicx}
\usepackage{natbib}
\usepackage{multirow}
\usepackage{amsmath}
\usepackage{amssymb}
\usepackage{bbm}
\usepackage{bm}
\usepackage{booktabs}
\usepackage{array} 
\usepackage{blindtext}
\usepackage{comment}

\newcommand{\tableCellHeight}{1.1}
\newcommand{\tabstyle}[1]{
  \setlength{\tabcolsep}{#1}
  \renewcommand{\arraystretch}{\tableCellHeight}
  \centering
  \footnotesize
}

\usepackage{pifont}

\newcommand{\romannum}[1]{\romannumeral #1} 

\usepackage[dvipsnames]{xcolor}
\usepackage{color, colortbl}
\definecolor{citecolor}{HTML}{0071bc}
\definecolor{tabhighlight}{HTML}{e5e5e5}
\usepackage[colorlinks,citecolor=citecolor]{hyperref}
\newcommand{\hgreen}[1]{\textcolor{ForestGreen}{\textbf{#1}}} 
\newcommand{\hred}[1]{\textcolor{WildStrawberry}{\textbf{#1}}} 

\makeatletter
\renewcommand\paragraph{
  \@startsection{paragraph} 
  {4} 
  {\z@} 
  {.5em \@plus1ex \@minus.2ex} 
  {-.5em} 
  {\normalfont\normalsize\bfseries} 
}
\makeatother
\begin{document}
\sloppy

\title{Learning to Prompt for Vision-Language Models 
}


\author{Kaiyang Zhou         \and
        Jingkang Yang \and
        Chen Change Loy \and
        Ziwei Liu
}


\institute{Kaiyang Zhou \at
              S-Lab, Nanyang Technological University, Singapore \\
              \email{kaiyang.zhou@ntu.edu.sg}
           \and
           Jingkang Yang \at
              S-Lab, Nanyang Technological University, Singapore \\
              \email{jingkang001@ntu.edu.sg}
           \and
           Chen Change Loy \at
              S-Lab, Nanyang Technological University, Singapore \\
              \email{ccloy@ntu.edu.sg}
           \and
           Ziwei Liu \at
           S-Lab, Nanyang Technological University, Singapore \\
           \email{ziwei.liu@ntu.edu.sg}
}

\date{Received: date / Accepted: date}

\maketitle

\begin{abstract}
Large pre-trained vision-language models like CLIP have shown great potential in learning representations that are transferable across a wide range of downstream tasks. Different from the traditional representation learning that is based mostly on discretized labels, vision-language pre-training aligns images and texts in a common feature space, which allows zero-shot transfer to a downstream task via \emph{prompting}, i.e., classification weights are synthesized from natural language describing classes of interest. In this work, we show that a major challenge for deploying such models in practice is prompt engineering, which requires domain expertise and is extremely time-consuming---one needs to spend a significant amount of time on words tuning since a slight change in wording could have a huge impact on performance. Inspired by recent advances in prompt learning research in natural language processing (NLP), we propose \emph{Context Optimization (CoOp)}, a simple approach specifically for adapting CLIP-like vision-language models for downstream image recognition. Concretely, CoOp models a prompt's context words with learnable vectors while the entire pre-trained parameters are kept fixed. To handle different image recognition tasks, we provide two implementations of CoOp: unified context and class-specific context. Through extensive experiments on 11 datasets, we demonstrate that CoOp requires as few as one or two shots to beat hand-crafted prompts with a decent margin and is able to gain significant improvements over prompt engineering with more shots, e.g., with 16 shots the average gain is around 15\% (with the highest reaching over 45\%). Despite being a learning-based approach, CoOp achieves superb domain generalization performance compared with the zero-shot model using hand-crafted prompts.
\end{abstract}

\section{Introduction} \label{sec:intro}

\begin{figure*}[t]
    \centering
    \includegraphics[width=.9\textwidth]{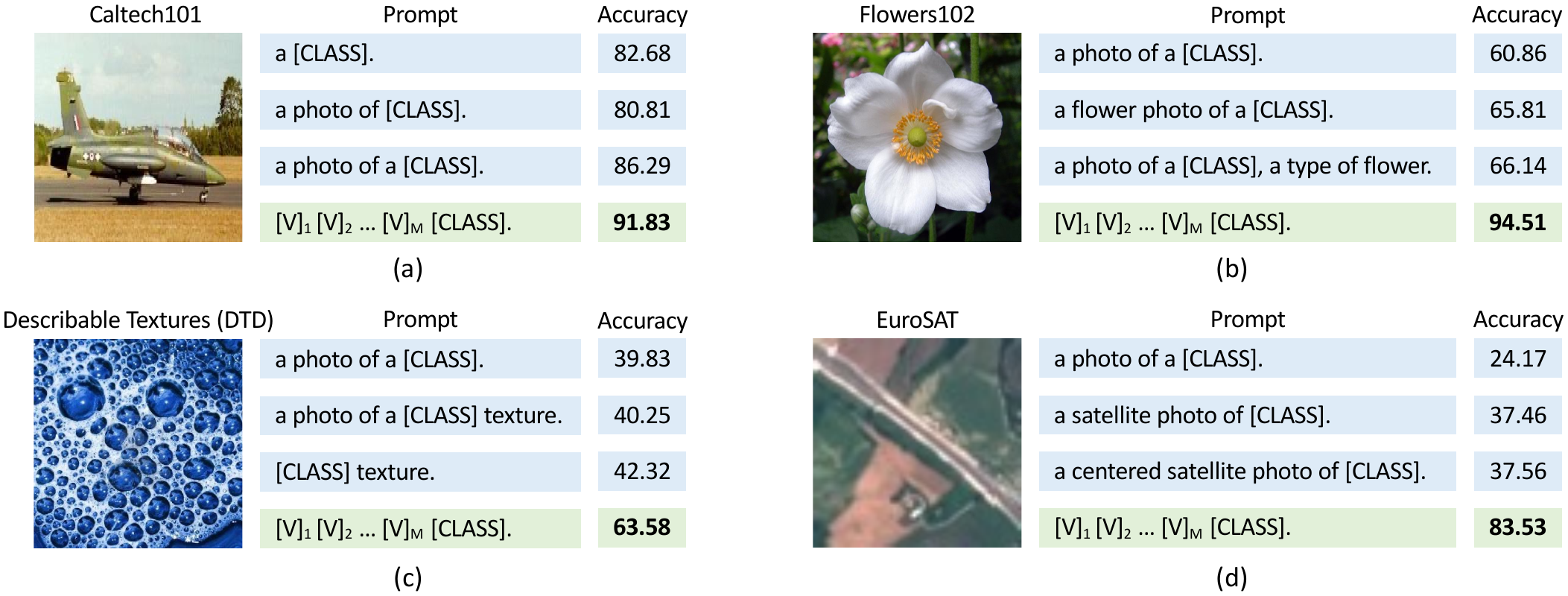}
    \caption{\textbf{Prompt engineering vs Context Optimization (CoOp)}. The former needs to use a held-out validation set for words tuning, which is inefficient; the latter automates the process and requires only a few labeled images for learning.
    }
    \label{fig:motivation}
\end{figure*}

A common approach for building state-of-the-art visual recognition systems is to train vision models to predict for a fixed set of object categories using discrete labels~\citep{he2016deep,dosovitskiy2021image}. From a technical point of view, this is achieved by matching image features---produced by a vision model like ResNet~\citep{he2016deep} or ViT~\citep{dosovitskiy2021image}---with a fixed set of weights that are seen as visual concepts and initialized randomly. Although training categories often have a textual form, such as ``goldfish'' or ``toilet paper,'' they will be converted into discrete labels just for easing the computation of the cross-entropy loss, leaving the semantics encapsulated in texts largely unexploited. Such a learning paradigm limits visual recognition systems to closed-set visual concepts, making them unable to deal with new categories since additional data are required for learning a new classifier.

Recently, vision-language pre-training such as CLIP~\citep{radford2021learning} and ALIGN~\citep{jia2021scaling} has emerged as a promising alternative for visual representation learning. The main idea is to align images and raw texts using two separate encoders---one for each modality. For instance, both CLIP and ALIGN formulate the learning objective as a contrastive loss, which pulls together images and their textual descriptions while pushes away unmatched pairs in the feature space. By pre-training at a large scale, models can learn diverse visual concepts and can readily be transferred to any downstream task through \emph{prompting}~\citep{radford2021learning,jia2021scaling,furst2021cloob,li2021supervision,singh2021flava,yuan2021florence}. In particular, for any new classification task one can first synthesize the classification weights by giving sentences describing task-relevant categories to the text encoder, and then compare with image features produced by the image encoder.

We observe that for pre-trained vision-language models, the text input, known as prompt, plays a key role in downstream datasets. However, identifying the right prompt is a non-trivial task, which often takes a significant amount of time for words tuning---a slight change in wording could make a huge difference in performance. For instance, for Caltech101 (Figure~\ref{fig:motivation}(a), 2nd vs 3rd prompt), adding ``a'' before the class token brings more than 5\% increase in accuracy. Moreover, prompt engineering also requires prior knowledge about the task and ideally the language model's underlying mechanism. This is exemplified in Figure~\ref{fig:motivation}(b-d) where adding task-relevant context can lead to significant improvements, i.e., ``flower'' for Flowers102, ``texture'' for DTD and ``satellite'' for EuroSAT. Tuning the sentence structure could bring further improvements, e.g., putting ``a type of flower'' after the class token for Flowers102, keeping only ``texture'' in the context for DTD, and adding ``centered'' before ``satellite photo'' for EuroSAT. However, even with extensive tuning, the resulting prompts are by no means guaranteed to be optimal for these downstream tasks.

\begin{figure*}[t]
    \centering
    \includegraphics[width=.9\textwidth]{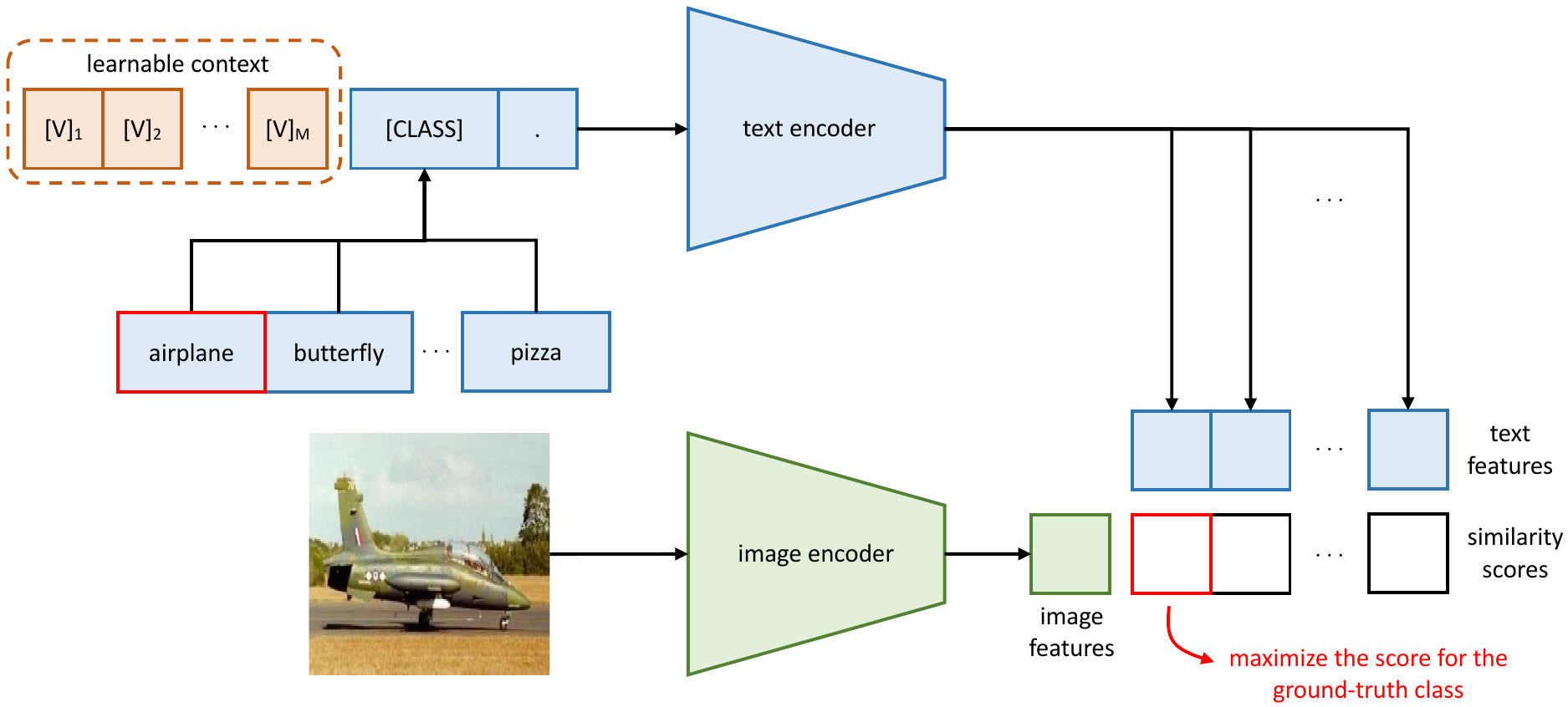}
    \caption{\textbf{Overview of Context Optimization (CoOp)}. The main idea is to model a prompt's context using a set of learnable vectors, which can be optimized through minimizing the classification loss. Two designs are proposed: one is unified context, which shares the same context vectors with all classes; and the other is class-specific context, which learns for each class a specific set of context vectors.}
    \label{fig:method}
\end{figure*}

Inspired by recent prompt learning research in natural language processing (NLP)~\citep{shin2020autoprompt,jiang2020can,zhong2021factual}, we propose a simple approach called \emph{Context Optimization (CoOp)}\footnote{CoOp is pronounced as /ku:p/.} to automate prompt engineering, specifically for pre-trained vision-language models. Concretely, CoOp models a prompt's context words with learnable vectors, which could be initialized with either random values or pre-trained word embeddings (see Figure~\ref{fig:method}). Two implementations are provided to handle tasks of different natures: one is based on unified context, which shares the same context with all classes and works well on most categories; while the other is based on class-specific context, which learns a specific set of context tokens for each class and is found to be more suitable for some fine-grained categories. During training, we simply minimize prediction errors using the cross-entropy loss with respect to the learnable context vectors while keeping the entire pre-trained parameters fixed. The gradients can be back-propagated all the way through the text encoder, distilling the rich knowledge encoded in the parameters for learning task-relevant context.

To demonstrate the effectiveness of CoOp, we benchmark on 11 datasets, which cover a diverse set of visual recognition tasks including classification on generic objects, scenes, actions and fine-grained categories, as well as specialized tasks like recognizing textures and satellite imagery. The results show that CoOp effectively turns pre-trained vision-language models into data-efficient visual learners, requiring as few as one or two shots to beat hand-crafted prompts with a decent margin. The performance can be further boosted by using more shots, e.g., with 16 shots the margin over hand-crafted prompts averages at around 15\% and reaches over 45\% for the highest. CoOp also outperforms the linear probe model, which is known as a strong few-shot learning baseline~\citep{tian2020rethinking}. Furthermore, CoOp demonstrates much stronger robustness than the zero-shot model (which uses manual prompts) to domain shifts, despite being a learning-based approach.

In summary, we make the following contributions:
\begin{enumerate}
	\item We present a timely study on the adaptation of recently proposed vision-language models in downstream applications and identify a critical problem associated with the deployment efficiency, i.e., prompt engineering.
	\item To automate prompt engineering specifically for pre-trained vision-language models, we propose a simple approach based on continuous prompt learning and provide two implementations that can handle different recognition tasks.
	\item We for the first time show that the proposed prompt learning-based approach outperforms both hand-crafted prompts and the linear probe model in terms of downstream transfer learning performance and robustness under domain shifts for large vision-language models.
	\item We open-source our project at \url{https://github.com/KaiyangZhou/CoOp}.
\end{enumerate}

We hope the findings together with the open-source code can inspire and facilitate future research on efficient adaptation methods for large vision-language models---an emerging topic related to democratization of foundation models~\citep{bommasani2021opportunities} i.e., making them easier and cheaper to adapt for the wider community.

\section{Related Work} \label{sec:related_work}

\subsection{Vision-Language Models} \label{sec:related_work;subsec:vl_models}
Vision-language models have recently demonstrated great potential in learning generic visual representations and allowing zero-shot transfer to a variety of downstream classification tasks via prompting~\citep{radford2021learning,jia2021scaling,zhang2020contrastive,singh2021flava,yuan2021florence}.

To our knowledge, the recent developments in vision-language learning, particularly CLIP~\citep{radford2021learning} and ALIGN~\citep{jia2021scaling}, are largely driven by advances in the following three areas: i) text representation learning with Transformers~\citep{vaswani2017attention}, ii) large-minibatch contrastive representation learning~\citep{chen2020simple,he2020momentum,henaff2020data}, and iii) web-scale training datasets---CLIP benefits from 400 million curated image-text pairs while ALIGN exploits 1.8 billion noisy image-text pairs.

The idea of mapping images and text onto a common embedding space has been studied since nearly a decade ago~\citep{socher2013zero,frome2013devise,elhoseiny2013write}, but with drastically different technologies. For text features extraction, early work has mainly utilized pre-trained word vectors~\citep{socher2013zero,frome2013devise} or the hand-crafted TF-IDF features~\citep{elhoseiny2013write,lei2015predicting}. Matching images and text features has been formulated as metric learning~\citep{frome2013devise}, multi-label classification~\citep{joulin2016learning,gomez2017self}, n-gram language learning~\citep{li2017learning}, and the recently proposed captioning~\citep{desai2021virtex}.

Our work is orthogonal to recent research in vision-language models, aiming to facilitate the adaptation and deployment of such models in downstream datasets.

\subsection{Prompt Learning in NLP} \label{sec:related_work;subsec:prompt_nlp}
Knowledge probing for large pre-trained language models, formally defined by \citet{petroni2019language} as ``fill-in-the-blank'' cloze tests, has recently sparked interest in prompt learning research in NLP~\citep{shin2020autoprompt,jiang2020can,li2021prefix,zhong2021factual,lester2021power,gao2020making,liu2021gpt}.

The basic idea of knowledge probing is to induce pre-trained language models to generate answers given cloze-style prompts, which can benefit a number of downstream tasks, such as sentiment analysis. \citet{jiang2020can} propose to generate candidate prompts through text mining and paraphrasing, and identify the optimal ones that give the highest training accuracy. \citet{shin2020autoprompt} introduce a gradient-based approach, which searches for tokens with the largest gradient changes in the label likelihood.

Most related to our work are continuous prompt learning methods~\citep{zhong2021factual,li2021prefix,lester2021power} which optimize continuous vectors in the word embedding space. A drawback of such methods compared to searching discrete tokens is the lack of a clear way to visualize what ``words'' are learned for the vectors. We refer readers to \citet{liu2021pre} for a comprehensive survey in the topic of prompt learning in NLP.

It is worth noting that we are the first to apply prompt learning to the adaptation of large vision-language models in computer vision---which we view as an important topic for democratizing foundation models~\citep{bommasani2021opportunities}---and justify that prompt learning not only brings significant improvements to computer vision tasks in terms of transfer learning performance but also produces robust models that can handle domain shifts.

\section{Methodology} \label{sec:method}

\subsection{Vision-Language Pre-training}
We briefly introduce vision-language pre-training with a particular focus on CLIP~\citep{radford2021learning}. Our approach is applicable to broader CLIP-like vision-language models.

\paragraph{Models}
CLIP consists of two encoders, one for images and the other for text. The image encoder aims to map high-dimensional images into a low-dimensional embedding space. The architecture of the image encoder can take the form of a CNN like ResNet-50~\citep{he2016deep} or a ViT~\citep{dosovitskiy2021image}. On the other hand, the text encoder is built on top of a Transformer~\citep{vaswani2017attention} and aims to generate text representations from natural language.

Specifically, given a sequence of words (tokens), such as ``a photo of a dog,'' CLIP first converts each one of the token (including punctuation) into a lower-cased byte pair encoding (BPE) representation~\citep{sennrich2016neural}, which is essentially a unique numeric ID. The vocabulary size in CLIP is 49,152. To facilitate minibatch processing, each text sequence is encompassed with the \texttt{[SOS]} and \texttt{[EOS]} tokens and capped at a fixed length of 77. After that, the IDs are mapped to 512-D word embedding vectors, which are then passed on to the Transformer. Finally, the features at the \texttt{[EOS]} token position are layer normalized and further processed by a linear projection layer.

\paragraph{Training}
CLIP is trained to align the two embedding spaces learned for images and text respectively. Specifically, the learning objective is formulated as a contrastive loss. Given a batch of image-text pairs, CLIP maximizes the cosine similarity for matched pairs while minimizes the cosine similarity for all other unmatched pairs. To learn diverse visual concepts that are more transferable to downstream tasks, CLIP's team collects a large training dataset consisting of 400 million image-text pairs.

\paragraph{Zero-Shot Inference}
Since CLIP is pre-trained to predict whether an image matches a textual description, it naturally fits zero-shot recognition. This is achieved by comparing image features with the classification weights synthesized by the text encoder, which takes as input textual descriptions specifying classes of interest. Formally, let $\bm{f}$ be image features extracted by the image encoder for an image $\bm{x}$ and $\{ \bm{w}_i \}_{i=1}^K$ a set of weight vectors generated by the text encoder. $K$ denotes the number of classes and each $\bm{w}_i$ is derived from a prompt that could have the form of ``a photo of a [CLASS].''~where the class token is replaced by the specific class name, such as ``cat,'' ``dog'' or ``car.'' The prediction probability is then computed as
\begin{equation} \label{eq:prob}
p(y=i|\bm{x}) = \frac{\exp( \cos(\bm{w_i}, \bm{f})/\tau) }{\sum_{j=1}^K \exp( \cos(\bm{w_j}, \bm{f}) /\tau)},
\end{equation}
where $\tau$ is a temperature parameter learned by CLIP and $\cos(\cdot, \cdot)$ denotes cosine similarity.

Compared with the traditional classifier learning approach where closed-set visual concepts are learned from random vectors, vision-language pre-training allows open-set visual concepts to be explored through a high-capacity text encoder, leading to a broader semantic space and in turn making the learned representations more transferable to downstream tasks.

\subsection{Context Optimization}
We propose Context Optimization (CoOp), which avoids manual prompt tuning by modeling context words with continuous vectors that are end-to-end learned from data while the massive pre-trained parameters are frozen. An overview is shown in Figure~\ref{fig:method}. Below we provide several different implementations.

\paragraph{Unified Context}
We first introduce the unified context version, which shares the same context with all classes. Specifically, the prompt given to the text encoder $g(\cdot)$ is designed with the following form,
\begin{equation} \label{eq:prompt_cls_end}
\bm{t} = [\text{V}]_1 [\text{V}]_2 \hdots [\text{V}]_M [\text{CLASS}],
\end{equation}
where each $[\text{V}]_m$ ($m\!\in\!\{1, \hdots, M\}$) is a vector with the same dimension as word embeddings (i.e., 512 for CLIP), and $M$ is a hyperparameter specifying the number of context tokens.

By forwarding a prompt $\bm{t}$ to the text encoder $g(\cdot)$, we can obtain a classification weight vector representing a visual concept (still from the \texttt{[EOS]} token position). The prediction probability is computed as
\begin{equation} \label{eq:prob_coop}
p(y=i|\bm{x}) = \frac{\exp( \cos(g(\bm{t}_i), \bm{f}) /\tau)}{\sum_{j=1}^K \exp( \cos(g(\bm{t}_j), \bm{f}) /\tau)},
\end{equation}
where the class token within each prompt $\bm{t}_i$ is replaced by the corresponding word embedding vector(s) of the $i$-th class name.

Other than placing the class token at the end of a sequence as in Equation~(\ref{eq:prompt_cls_end}), we can also put it in the middle like
\begin{equation} \label{eq:prompt_cls_mid}
\bm{t} = [\text{V}]_1 \hdots [\text{V}]_{\frac{M}{2}} [\text{CLASS}] [\text{V}]_{\frac{M}{2}+1} \hdots [\text{V}]_M,
\end{equation}
which increases flexibility for learning---the prompt is allowed to either fill the latter cells with supplementary descriptions or cut off the sentence earlier by using a termination signal such as full stop.

\paragraph{Class-Specific Context}
Another option is to design class-specific context (CSC) where context vectors are independent to each class, i.e., $[\text{V}]^i_1 [\text{V}]^i_2 \hdots [\text{V}]^i_M \neq [\text{V}]^j_1 [\text{V}]^j_2 \hdots [\text{V}]^j_M$ for $i\!\neq\!j$ and $i, j \in \{1, \hdots, K\}$. As an alternative to unified context, we find that CSC is particularly useful for some fine-grained classification tasks.

\paragraph{Training}
is performed to minimize the standard classification loss based on the cross-entropy, and the gradients can be back-propagated all the way through the text encoder $g(\cdot)$, making use of the rich knowledge encoded in the parameters to optimize the context. The design of continuous representations also allows full exploration in the word embedding space, which facilitates the learning of task-relevant context.

\subsection{Discussion}
Our approach specifically addresses the emerging problem of the adaptation of recently proposed large vision-language models such as CLIP~\citep{radford2021learning}. There are some differences that distinguish our approach from the prompt learning methods developed in NLP for language models (e.g., GPT-3~\citep{brown2020language}). First, the backbone architectures are clearly different for CLIP-like models and language models---the former take both visual and textual data as input and produce alignment scores used for image recognition, while the latter are tailored to handle textual data only. Second, the pre-training objectives are different: contrastive learning vs autoregressive learning. This would lead to different model behaviors and thus require different module designs.

\begin{figure*}[t]
    \centering
    \includegraphics[width=.98\textwidth]{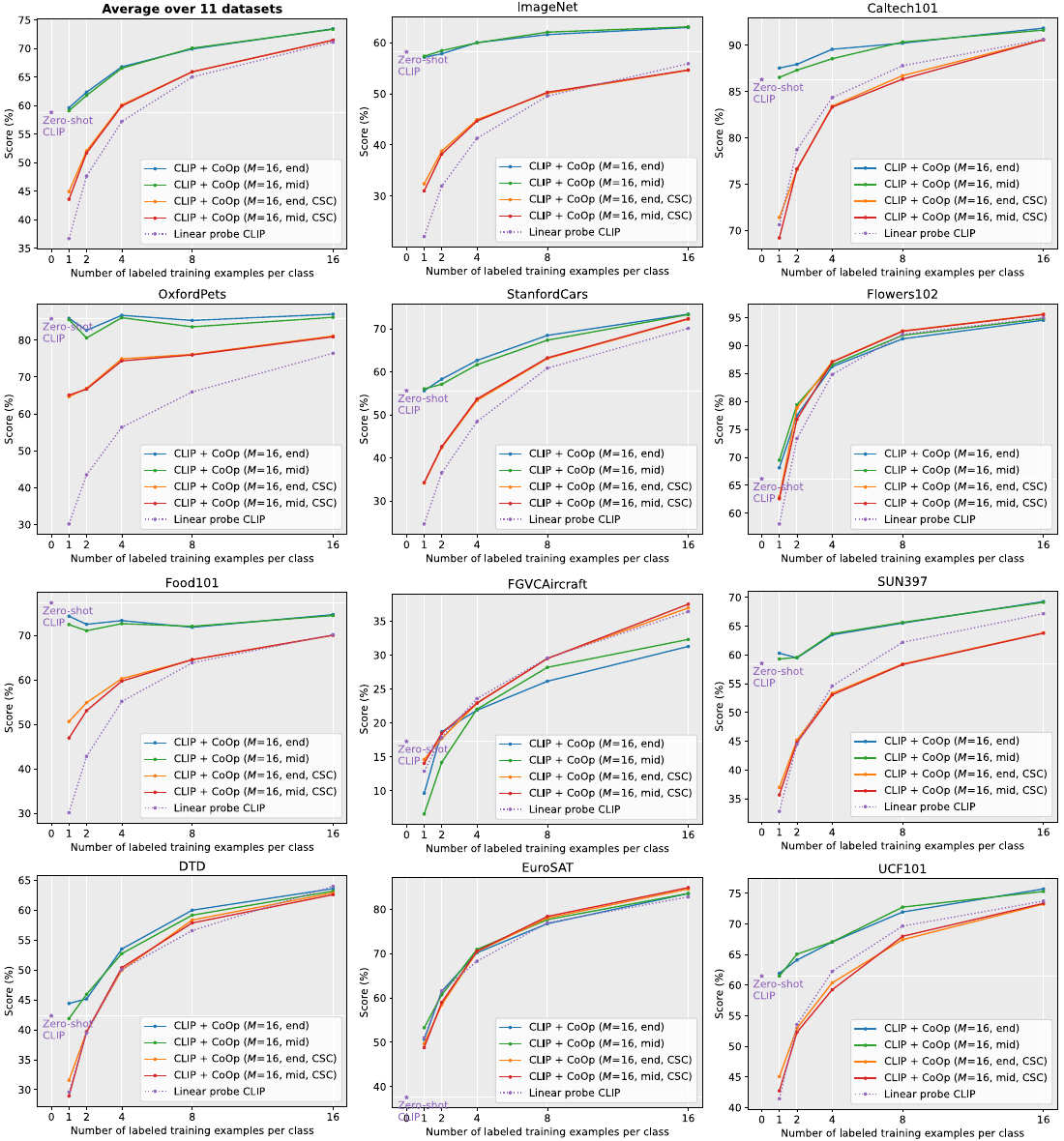}
    \caption{\textbf{Main results of few-shot learning on the 11 datasets}. Overall, CoOp effectively turns CLIP into a strong few-shot learner (solid lines), achieving significant improvements over zero-shot CLIP (stars) and performing favorably against the linear probe alternative (dashed lines). $M$ denotes the context length. ``end'' or ``mid'' means putting the class token in the end or middle. CSC means class-specific context.}
    \label{fig:main_results}
\end{figure*}

\section{Experiments} \label{sec:exp}

\subsection{Few-Shot Learning} \label{sec:exp;subsec:fewshot}
\paragraph{Datasets}
We select 11 publicly available image classification datasets used in CLIP: ImageNet~\citep{deng2009imagenet}, Caltech101~\citep{fei2004learning}, OxfordPets~\citep{parkhi2012cats}, StanfordCars~\citep{krause20133d}, Flowers102~\citep{nilsback2008automated}, Food101~\citep{bossard2014food}, FGVCAircraft~\citep{maji2013fine}, SUN397~\citep{xiao2010sun}, DTD~\citep{cimpoi2014describing}, EuroSAT~\citep{helber2019eurosat} and UCF101~\citep{soomro2012ucf101} (see Appendix~\ref{appx:datasets_details} for their statistics). These datasets constitute a comprehensive benchmark, which covers a diverse set of vision tasks including classification on generic objects, scenes, actions and fine-grained categories, as well as specialized tasks like recognizing textures and satellite imagery. We follow the few-shot evaluation protocol adopted in CLIP~\citep{radford2021learning}, using 1, 2, 4, 8 and 16 shots for training respectively and deploying models in the full test sets. The average results over three runs are reported for comparison.

\paragraph{Training Details}
CoOp has four versions: positioning the class token in the end or middle; unified context vs CSC. Unless otherwise stated, ResNet-50~\citep{he2016deep} is used as the image encoder's backbone and the number of context tokens $M$ is set to 16. Investigations on other design choices are discussed in Section~\ref{sec:exp;subsec:analysis}. All models are built on top of CLIP's open-source code.\footnote{\url{https://github.com/openai/CLIP}.} CoOp's context vectors are randomly initialized by drawing from a zero-mean Gaussian distribution with standard deviation equal to 0.02. Training is done with SGD and an initial learning rate of 0.002, which is decayed by the cosine annealing rule. The maximum epoch is set to 200 for 16/8 shots, 100 for 4/2 shots, and 50 for 1 shot (except for ImageNet where the maximum epoch is fixed to 50). To mitigate explosive gradients observed in the early training iterations, we use the warmup trick by fixing the learning rate to $1e\!-\!5$, only during the first epoch.

\paragraph{Baseline Methods}
We compare CoOp with two baseline methods. The first is zero-shot CLIP, which is based on hand-crafted prompts. We follow the guideline of prompt engineering introduced by \citet{radford2021learning}. For generic objects and scenes, ``a photo of a [CLASS].''~is adopted. For fine-grained categories, task-relevant context is added like ``a type of pet'' for OxfordPets and ``a type of food'' for Food101. When it comes to specialized tasks such as recognizing textures in DTD, the prompt is customized as ``[CLASS] texture.''~where the class names are adjectives like ``bubbly'' and ``dotted.'' See Appendix~\ref{appx:datasets_details} for the details. The second baseline is the linear probe model. As suggested by \citet{radford2021learning} and a recent study on few-shot learning~\citep{tian2020rethinking}, training a linear classifier on top of high-quality pre-trained models' features (like CLIP) can easily achieve performance that is on a par with that of state-of-the-art few-shot learning methods, which are often much more sophisticated. We follow the same training method used by \citet{radford2021learning} to train the linear probe model.

\paragraph{Comparison with Hand-Crafted Prompts}
Figure~\ref{fig:main_results} summarizes the results. Our default model is CLIP+CoOp with the class token positioned in the end. The two different ways of positioning the class token achieve similar performance as their curves highly overlap. From the average performance displayed in the top-left corner, we observe that CLIP+CoOp is a strong few-shot learner, requiring only two shots on average to obtain a decent margin over zero-shot CLIP. Given 16 shots for training, the average gap brought by CoOp can be further increased to around 15\%.

\begin{figure}[t]
	\centering
	\includegraphics[width=\columnwidth]{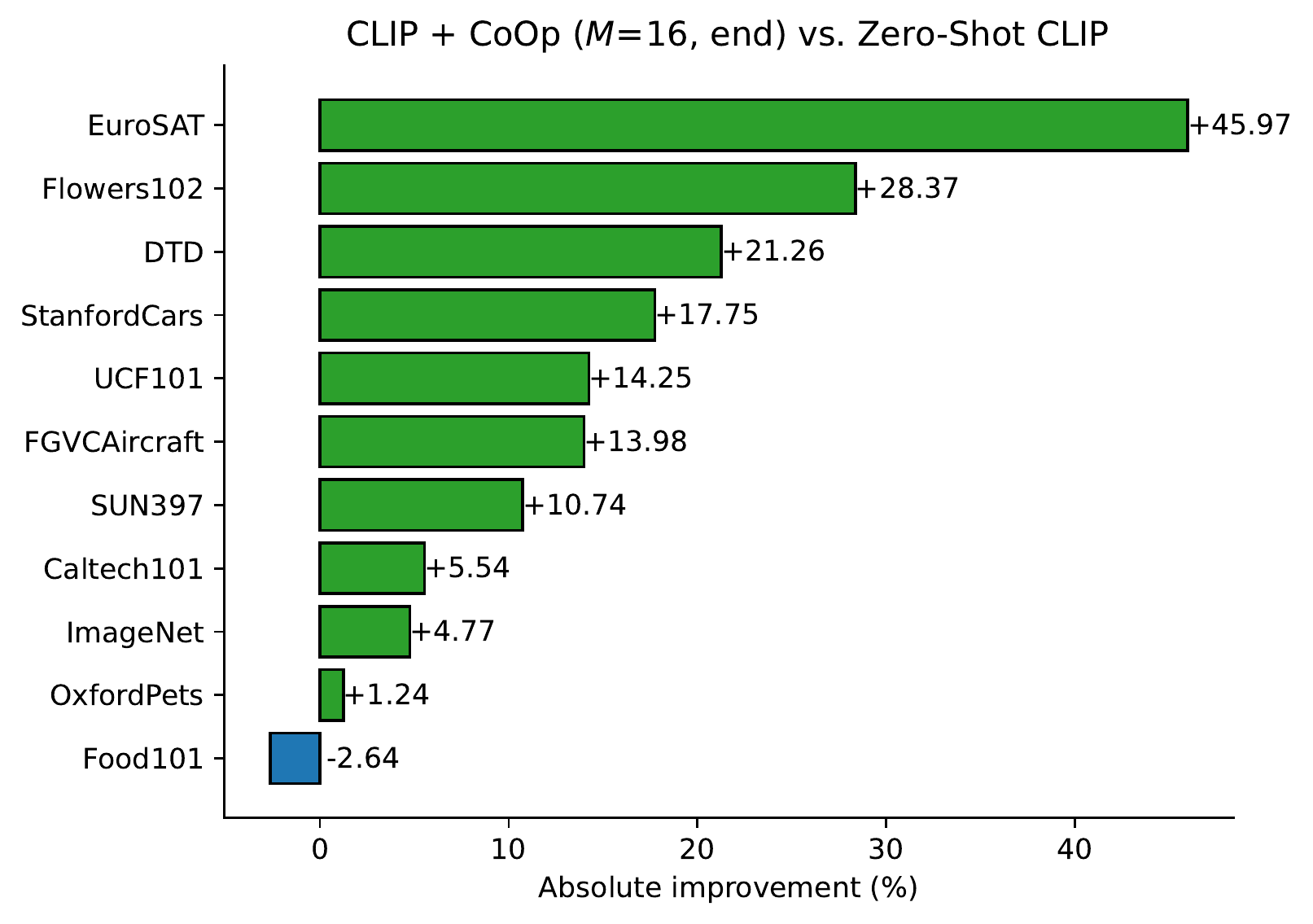}
	\caption{Comparison with hand-crafted prompts.}
	\label{fig:abs_improve}
\end{figure}

Figure~\ref{fig:abs_improve} ranks the absolute improvements obtained by CoOp at 16 shots over hand-crafted prompts. Huge improvements are observed on specialized tasks namely EuroSAT and DTD where the increase in performance reaches over 45\% and 20\% respectively. The jumps in performance are also significant (those more than 10\%) on most fine-grained datasets including Flowers102, StanfordCars and FGVCAircraft, as well as on scene and action recognition datasets (i.e., SUN397 \& UCF101). Since ImageNet is a challenging dataset that contains 1,000 classes, the 4.77\% improvement is also noteworthy. In contrast, the increases on the two fine-grained datasets, OxfordPets and Food101, are less appealing.\footnote{We find that the negative results on Food101, for learning-based models including CoOp and linear probe, are caused by the noisy training data with ``intense colors and sometimes wrong labels''~\citep{bossard2014food}.} By digging into CLIP+CoOp's curves on these two datasets in Figure~\ref{fig:main_results}, we find there is a loss of momentum in performance improvements even with more shots used, seemingly an overfitting problem. A potential solution is to impose higher regularization like increasing the weight decay. Nonetheless, the overall results are strong enough to serve as evidence of CoOp's capability of learning task-relevant prompts in a data-efficient manner.

\begin{table*}[t]
    \tabstyle{7pt}
    \caption{Comparison with zero-shot CLIP on robustness to distribution shift using different vision backbones. $M$: CoOp's context length.}
    \label{tab:robustness_various_archs}
    \begin{tabular}{l ccccc}
    \toprule
    & Source & \multicolumn{4}{c}{Target} \\ \cmidrule(lr){2-2} \cmidrule(lr){3-6}
    Method & ImageNet & -V2 & -Sketch & -A & -R \\
    \midrule
    \textbf{ResNet-50} & \\
    Zero-Shot CLIP & 58.18 & 51.34 & 33.32 & 21.65 & 56.00 \\
    Linear Probe CLIP & 55.87 & 45.97 & 19.07 & 12.74 & 34.86 \\
    \rowcolor{tabhighlight}
    CLIP + CoOp ($M\!=\!16$) & 62.95 & 55.11 & 32.74 & 22.12 & 54.96 \\
    \rowcolor{tabhighlight}
    CLIP + CoOp ($M\!=\!4$) & \textbf{63.33} & \textbf{55.40} & \textbf{34.67} & \textbf{23.06} & \textbf{56.60} \\
    \midrule
    \textbf{ResNet-101} & \\
    Zero-Shot CLIP & 61.62 & 54.81 & 38.71 & 28.05 & 64.38 \\
    Linear Probe CLIP & 59.75&	50.05&	26.80&	19.44&	47.19 \\
    \rowcolor{tabhighlight}
    CLIP + CoOp ($M\!=\!16$) & \textbf{66.60} & \textbf{58.66} & 39.08 & 28.89 & 63.00 \\
    \rowcolor{tabhighlight}
    CLIP + CoOp ($M\!=\!4$) & 65.98 & 58.60 & \textbf{40.40} & \textbf{29.60} & \textbf{64.98} \\
    \midrule
    \textbf{ViT-B/32} & \\
    Zero-Shot CLIP & 62.05 & 54.79 & 40.82 & 29.57 & \textbf{65.99} \\
    Linear Probe CLIP & 59.58&	49.73&	28.06&	19.67&	47.20\\
    \rowcolor{tabhighlight}
    CLIP + CoOp ($M\!=\!16$) & \textbf{66.85} & 58.08 & 40.44 & 30.62 & 64.45 \\
    \rowcolor{tabhighlight}
    CLIP + CoOp ($M\!=\!4$) & 66.34 & \textbf{58.24} & \textbf{41.48} & \textbf{31.34} & 65.78 \\
    \midrule
    \textbf{ViT-B/16} & \\
    Zero-Shot CLIP & 66.73 & 60.83 & 46.15 & 47.77 & 73.96 \\
    Linear Probe CLIP & 65.85&	56.26&	34.77&	35.68&	58.43\\
    \rowcolor{tabhighlight}
    CLIP + CoOp ($M\!=\!16$) & \textbf{71.92} & 64.18 & 46.71 & 48.41 & 74.32 \\
    \rowcolor{tabhighlight}
    CLIP + CoOp ($M\!=\!4$) & 71.73 & \textbf{64.56} & \textbf{47.89} & \textbf{49.93} & \textbf{75.14} \\
    \bottomrule
    \end{tabular}
\end{table*}

\paragraph{Comparison with Linear Probe CLIP}
In terms of the overall performance (Figure~\ref{fig:main_results}, top-left), CLIP+CoOp demonstrates clear advantages over the linear probe model. The latter requires more than 4 shots on average to match the zero-shot's performance while CoOp's average gain at 4 shots is already impressive. It is also clear that the gaps in the extreme low-data regime such as one or two shots are much larger, suggesting that CoOp is much more effective than learning a linear classifier from scratch for few-shot learning. We also observe that the linear probe model is comparable to CLIP+CoOp on the two specialized tasks (DTD \& EuroSAT) as well as on a couple of fine-grained datasets (Flowers102 \& FGVCAircraft)---this is not too surprising as the pre-trained CLIP space has been proved powerful, making the linear probe model a strong competitor. Nevertheless, CoOp's CSC version can beat the linear probe CLIP on the aforementioned datasets, and moreover, shows much better potential when more shots become available. We later show that CoOp obtains much stronger performance than the linear probe model in domain generalization.

\paragraph{Unified vs Class-Specific Context}
On average, using unified context leads to better performance. In terms of when to apply CSC and when not to, we have the following suggestions. For generic objects (ImageNet \& Caltech101), scenes (SUN397) and actions (UCF101), using unified context is clearly better. Unified context also works better on some fine-grained datasets including OxfordPets and Food101, but on others like StanfordCars, Flowers102 and FGVCAircraft the CSC version is preferred. CSC also yields better performance on the two specialized tasks, DTD and EuroSAT, at 16 shots in particular. However, CSC mostly underperforms unified context in challenging low-data scenarios (fewer than 8 shots), which makes sense because CSC has more parameters than unified context and needs more data for training.

\begin{figure*}[t]
    \centering
    \includegraphics[width=.8\textwidth]{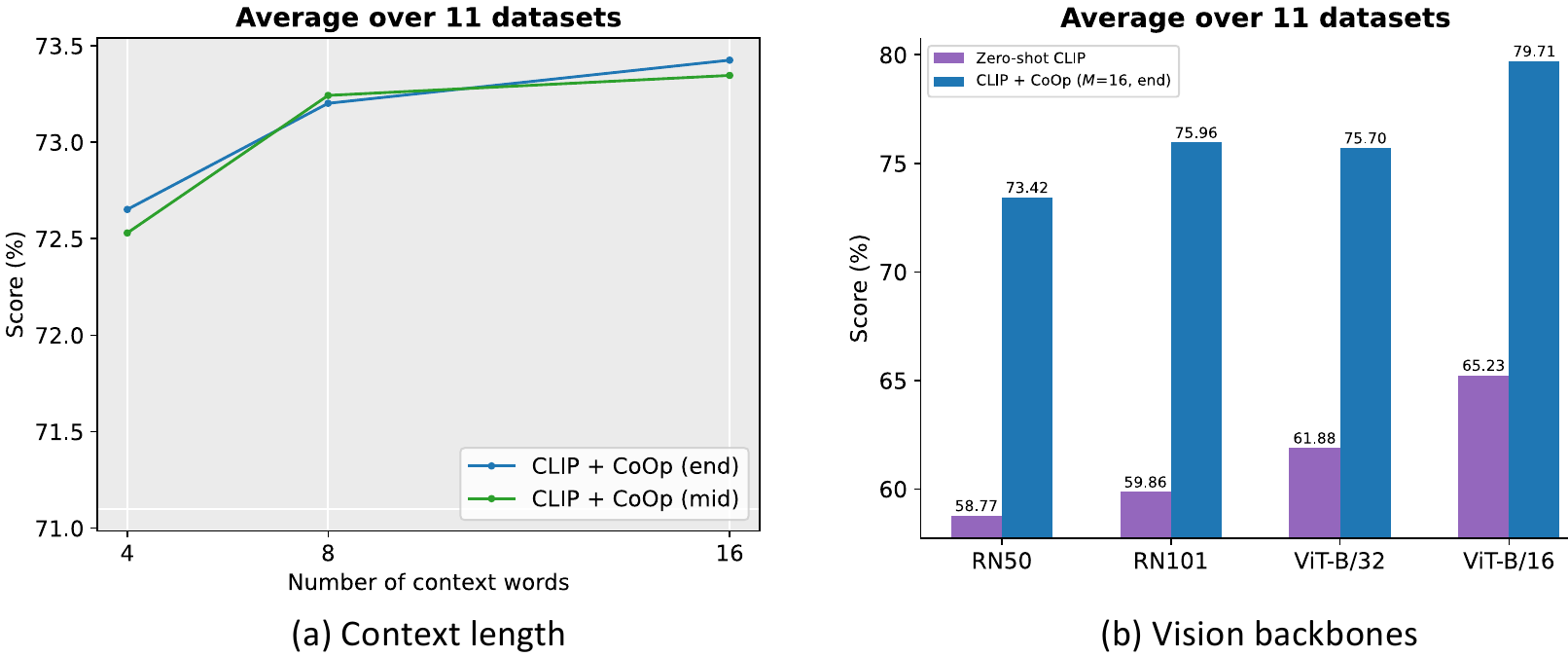}
    \caption{Investigations on CoOp's context length and various vision backbones.}
    \label{fig:study_ctxlen_visarch}
\end{figure*}

\begin{table*}[t]
    \tabstyle{8pt}
    \caption{Comparison with prompt engineering and prompt ensembling on ImageNet using different vision backbones.}
    \label{tab:vs_prompt_esb}
    \begin{tabular}{l cccc}
    \toprule
    Method & ResNet-50 & ResNet-101 & ViT-B/32 & ViT-B/16 \\
    \midrule
    Prompt engineering & 58.18 & 61.26 & 62.05 & 66.73 \\
    Prompt ensembling & 60.41 & 62.54 & 63.71 & 68.74 \\
    \rowcolor{tabhighlight}
    CoOp & \textbf{62.95} & \textbf{66.60} & \textbf{66.85} & \textbf{71.92} \\
    \bottomrule
    \end{tabular}
\end{table*}

\subsection{Domain Generalization} \label{sec:exp;subsec:dg}
Since CoOp requires training on a specific data distribution, it risks learning spurious correlations that are detrimental to generalization in unseen distributions (domains), as suggested in recent studies~\citep{taori2020measuring,zhou2021domain}. On the contrary, zero-shot CLIP is not tied to a specific data distribution and has exhibited strong robustness to distribution shifts~\citep{radford2021learning}. In this section, we aim to unveil how robust CoOp is to distribution shifts, in comparison to zero-shot CLIP and the linear probe model.

\paragraph{Datasets}
The source dataset is ImageNet. The target datasets are ImageNetV2~\citep{recht2019imagenet}, ImageNet-Sketch~\citep{wang2019learning}, ImageNet-A~\citep{hendrycks2021natural} and ImageNet-R~\citep{hendrycks2021many}, all of which have compatible class names with ImageNet allowing seamless transfer for the prompts learned by CoOp. ImageNetV2 is a reproduced test set using different sources while following ImageNet's data collection process. ImageNet-Sketch contains sketch images belonging to the same 1,000 ImageNet classes. Both ImageNet-A and -R contain 200 classes derived from a subset of ImageNet's 1,000 classes. The former consists of real-world adversarially filtered images that cause current ImageNet classifiers to produce low results, whereas the latter features a rendition of the ImageNet classes in diverse image styles such as paintings, cartoons and sculptures.

\paragraph{Results}
Table~\ref{tab:robustness_various_archs} summarizes the results (with a variety of vision backbones). It is surprising that CoOp enhances CLIP's robustness to distribution shifts, \emph{despite the exposure to the source dataset}. This suggests that the learned prompts are also generalizable. Moreover, it is interesting to see that using fewer context tokens leads to better robustness. In contrast, the linear probe model obtains much worse results on these target datasets, exposing its weakness in domain generalization. In Appendix~\ref{appx:dosco_2k}, we provide the domain generalization results on DOSCO-2k~\citep{zhou2022device}, a recently proposed benchmark focusing on contextual domain shift.

\subsection{Further Analysis} \label{sec:exp;subsec:analysis}

\begin{table}[t]
    \tabstyle{5pt}
    \caption{Random vs manual initialization.}
    \label{tab:init}
    \begin{tabular}{l c}
    \toprule
     & Avg \% \\
    \midrule
    $[\text{V}]_1 [\text{V}]_2 [\text{V}]_3 [\text{V}]_4$ & 72.65 \\
    ``a photo of a'' & 72.65 \\
    \bottomrule
    \end{tabular}
\end{table}

\paragraph{Context Length}
How many context tokens should be used? And is it better to have more context tokens? The results in Section~\ref{sec:exp;subsec:dg} suggest having a shorter context length benefits domain generalization (probably due to less overfitting as fewer parameters are learned). Here we study this hyperparameter for source datasets. Specifically, we repeat experiments on the 11 datasets by varying the context length from 4 to 8 to 16. The average results are shown in Figure~\ref{fig:study_ctxlen_visarch}(a), which indicate that having more context tokens leads to better performance and that positioning the class token in the middle gains more momentum with longer context length. To sum up, there is no golden rule for selecting perfect context length since one needs to balance between performance and robustness to distribution shift.

\paragraph{Vision Backbones}
Figure~\ref{fig:study_ctxlen_visarch}(b) summarizes the results on the 11 datasets using a variety of vision backbones covering both CNNs and ViTs. The results are expected: the more advanced the backbone, the better the performance. The gap between CoOp and hand-crafted prompts is significant across all architectures.

\paragraph{Comparison with Prompt Ensembling}
The authors of CLIP~\citep{radford2021learning} have suggested that additional improvements can be obtained by ensembling over multiple zero-shot classifiers generated using different hand-crafted prompts, such as ``a photo of the large [CLASS].'', ``a bad photo of the [CLASS].''~and ``a origami [CLASS].'', which reflect a different scale, view and abstraction respectively for an image. We are interested to know whether the prompts learned by CoOp can still maintain advantages when compared with prompt ensembling. For fair comparison, we use the select prompts from~\citet{radford2021learning}, which have been extensively tuned on ImageNet, to construct the ensemble classifier. Table~\ref{tab:vs_prompt_esb} shows the comparison and justifies the superiority of CoOp. Given the potential of prompt ensembling, future work could investigate how to improve CoOp from the ensembling perspective.

\begin{table*}[t]
    \tabstyle{1pt}
    \caption{The nearest words for each of the 16 context vectors learned by CoOp, with their distances shown in parentheses. N/A means non-Latin characters.}
    \label{tab:vis_ctx_vec}
    \begin{tabular}{c |r|r|r|r|r}
    \toprule
    \# & ImageNet & Food101 & OxfordPets & DTD & UCF101 \\
    \midrule
    1 & potd (1.7136) & lc (0.6752) & tosc (2.5952) & boxed (0.9433) & meteorologist (1.5377) \\
    2 & that (1.4015) & enjoyed (0.5305) & judge (1.2635) & seed (1.0498) & exe (0.9807) \\
    3 & filmed (1.2275) & beh (0.5390) & fluffy (1.6099) & anna (0.8127) & parents (1.0654) \\
    4 & fruit (1.4864) & matches (0.5646) & cart (1.3958) & mountain (0.9509) & masterful (0.9528) \\
    5 & ,... (1.5863) & nytimes (0.6993) & harlan (2.2948) & eldest (0.7111) & fe (1.3574) \\
    6 & ° (1.7502) & prou (0.5905) & paw (1.3055) & pretty (0.8762) & thof (1.2841) \\
    7 & excluded (1.2355) & lower (0.5390) & incase (1.2215) & faces (0.7872) & where (0.9705) \\
    8 & cold (1.4654) & N/A & bie (1.5454) & honey (1.8414) & kristen (1.1921) \\
    9 & stery (1.6085) & minute (0.5672) & snuggle (1.1578) & series (1.6680) & imam (1.1297) \\
    10 & warri (1.3055) & $\sim$ (0.5529) & along (1.8298) & coca (1.5571) & near (0.8942) \\
    11 & marvelcomics (1.5638) & well (0.5659) & enjoyment (2.3495) & moon (1.2775) & tummy (1.4303) \\
    12 & .: (1.7387) & ends (0.6113) & jt (1.3726) & lh (1.0382) & hel (0.7644) \\
    13 & N/A & mis (0.5826) & improving (1.3198) & won (0.9314) & boop (1.0491) \\
    14 & lation (1.5015) & somethin (0.6041) & srsly (1.6759) & replied (1.1429) & N/A \\
    15 & muh (1.4985) & seminar (0.5274) & asteroid (1.3395) & sent (1.3173) & facial (1.4452) \\
    16 & .\# (1.9340) & N/A & N/A & piedmont (1.5198) & during (1.1755) \\
    \bottomrule
    \end{tabular}
\end{table*}

\paragraph{Comparison with Other Fine-tuning Methods}
We further compare CoOp with other fine-tuning methods: \romannum{1}) fine-tuning CLIP's image encoder; \romannum{2}) optimizing a transformation layer added to the text encoder's output; \romannum{3}) optimizing a bias term added to the text encoder's output. The results are shown in Table~\ref{tab:vs_other_finetune}. Obviously, fine-tuning the image encoder does not work well. Adding a transformation layer slightly improves upon the zero-shot model. Adding a bias term shows promising results, but still largely underperforms CoOp, which suggests that the gradients that went through the text encoder provide more useful information.

\begin{table}[t]
\tabstyle{3pt}
\caption{CoOp vs other fine-tuning methods on ImageNet (w/ 16 shots). $\Delta$: difference with the zero-shot model.}
\label{tab:vs_other_finetune}
\begin{tabular}{l cc}
\toprule
& ImageNet & $\Delta$ \\
\midrule
Zero-shot CLIP & 58.18 & - \\
Linear probe & 55.87 & \hred{-2.31} \\
Fine-tuning CLIP's image encoder & 18.28 & \hred{-39.90} \\
Optimizing transformation layer (text) & 58.86 & \hgreen{0.68} \\
Optimizing bias (text) & 60.93 & \hgreen{+2.75} \\
\rowcolor{tabhighlight}
CoOp & \textbf{62.95} & \hgreen{+4.77} \\
\bottomrule
\end{tabular}
\end{table}

\paragraph{Initialization}
We compare random initialization with manual initialization. The latter uses the embeddings of ``a photo of a'' to initialize the context vectors for the 11 datasets. For fair comparison, we also set the context length to 4 when using random initialization. Table~\ref{tab:init} suggests a ``good'' initialization does not make much difference. Though further tuning of the initialization words might help, in practice we suggest using the simple random initialization method.

\paragraph{Interpreting the Learned Prompts}
is difficult because the context vectors are optimized in a continuous space. We resort to an indirect way by searching within the vocabulary for words that are closest to the learned vectors based on the Euclidean distance. Note that CLIP~\citep{radford2021learning} uses the BPE representation~\citep{sennrich2016neural} for tokenization, so the vocabulary includes subwords that frequently appear in text, such as ``hu'' (subsumed by many words like ``hug'' and ``human''). Table~\ref{tab:vis_ctx_vec} shows the searched results on some datasets. We observe that a few words are somewhat relevant to the tasks, such as ``enjoyed'' for Food101, ``fluffy'' and ``paw'' for OxfordPets, and ``pretty'' for DTD. But when connecting all the nearest words together, the prompts do not make much sense. We also observe that when using manual initialization (like ``a photo of a''), the nearest words for the converged vectors are mostly the ones used for initialization. We conjecture that the learned vectors might encode meanings that are beyond the existing vocabulary. Overall, we are unable to draw any firm conclusion based on the observations because using nearest words to interpret the learned prompts could be inaccurate---the semantics of the vectors is not necessarily correlated with the nearest words.

\section{Conclusion, Limitations and Future Work}
Large pre-trained vision-language models have shown surprisingly powerful capabilities in diverse downstream applications. However, these models, also called vision foundation models given their ``critically central yet incomplete'' nature~\citep{bommasani2021opportunities}, need to be adapted using automated techniques for better downstream performance and efficiency.

Our research provides timely insights on how CLIP-like models can be turned into a data-efficient learner by using prompt learning, and reveals that despite being a learning-based approach, CoOp performs much better in domain generalization than manual prompts. The results serve as strong evidence that prompt learning has potential for large vision models. It is worth noting that our paper presents the first comprehensive study about adapting large vision models with prompt learning.

Though the performance is excellent, the results of CoOp are relatively difficult to interpret, like other continuous prompt learning methods in NLP. The experiments also reveal that CoOp is sensitive to noisy labels given the weak performance on Food101.

Nevertheless, the simplicity of CoOp allows easy extension for future work and there remain many interesting questions to explore, such as cross-dataset transfer~\citep{zhou2022conditional} and test-time adaptation~\citep{wang2020tent}. It would also be interesting to investigate more generic adaptation methods for mega-size vision models~\citep{jia2022visual,bahng2022visual,gao2021clip}. In summary, we hope the empirical findings and insights presented in this work could pave the way for future research on efficient adaptation methods for emerging foundation models, which is still a nascent research topic.

\begin{acknowledgements}
This work is supported by NTU NAP, MOE AcRF Tier 2 (T2EP20221-0033), and under the RIE2020 Industry Alignment Fund – Industry Collaboration Projects (IAF-ICP) Funding Initiative, as well as cash and in-kind contribution from the industry partner(s).
Corresponding author: Ziwei Liu (ziwei.liu@ntu.edu.sg).
\end{acknowledgements}

\begin{table*}[t]
    \tabstyle{6pt}
    \caption{Datasets statistics.}
    \label{tab:datasets}
    \begin{tabular}{l r r r r c}
    \toprule
    Dataset & Classes & Train & Val & Test & Hand-crafted prompt \\
    \midrule
    ImageNet & 1,000 & 1.28M & N/A & 50,000 & ``a photo of a [CLASS].'' \\
    Caltech101 & 100 & 4,128 & 1,649 & 2,465 & ``a photo of a [CLASS].'' \\ 
    OxfordPets & 37 & 2,944 & 736 & 3,669 & ``a photo of a [CLASS], a type of pet.'' \\
    StanfordCars & 196 & 6,509 & 1,635 & 8,041 & ``a photo of a [CLASS].'' \\
    Flowers102 & 102 & 4,093 & 1,633 & 2,463 & ``a photo of a [CLASS], a type of flower.'' \\
    Food101 & 101 & 50,500 & 20,200 & 30,300 & ``a photo of [CLASS], a type of food.'' \\
    FGVCAircraft & 100 & 3,334 & 3,333 & 3,333 & ``a photo of a [CLASS], a type of aircraft.'' \\
    SUN397 & 397 & 15,880 & 3,970 & 19,850 & ``a photo of a [CLASS].'' \\
    DTD & 47 & 2,820 & 1,128 & 1,692 & ``[CLASS] texture.'' \\
    EuroSAT & 10 & 13,500 & 5,400 & 8,100 & ``a centered satellite photo of [CLASS].'' \\
    UCF101 & 101 & 7,639 & 1,898 & 3,783 & ``a photo of a person doing [CLASS].'' \\
    \midrule
    ImageNetV2 & 1,000 & N/A & N/A & 10,000 & ``a photo of a [CLASS].'' \\
    ImageNet-Sketch & 1,000 & N/A & N/A & 50,889 & ``a photo of a [CLASS].'' \\
    ImageNet-A & 200 & N/A & N/A & 7,500 & ``a photo of a [CLASS].'' \\
    ImageNet-R & 200 & N/A & N/A & 30,000 & ``a photo of a [CLASS].'' \\
    \bottomrule
    \end{tabular}
\end{table*}

\appendix
\section*{Appendix}

\section{Datasets Details} \label{appx:datasets_details}
The detailed statistics of the 11 datasets, as well as the four variants of ImageNet, are shown in Table~\ref{tab:datasets}. The hand-crafted prompts used for zero-shot CLIP are also detailed in the table. For Caltech101, the ``BACKGROUND\_Google'' and ``Faces\_easy'' classes are discarded. For the video dataset, UCF101, the middle frame of each video is used as input to the image encoder.

\begin{table*}[t]
    \tabstyle{9pt}
    \caption{Domain generalization results on DOSCO-2k, a recently proposed benchmark focusing on broader contextual domain shift. Among the three approaches, CoOp and its follow-up, CoCoOp, contain learnable components while CLIP here denotes the zero-shot model. Both CoOp and CoCoOp use four learnable context tokens initialized with the word embeddings of ``a photo of a''. Bold denotes the best performance on each dataset for a specific architecture.}
    \label{tab:dosco_2k}
    \begin{tabular}{l cccccccc}
    \toprule
    & P-Air & P-Cars & P-Ctech & P-Ins & P-Mam & P-Pets & P-UCF & \textit{Avg} \\
    \midrule
    \textbf{ResNet-50} & \\
    CLIP & 16.1 & 56.1 & 86.7 & 62.7 & 59.7 & 84.0 & 60.6 & 60.9 \\
    CoOp & \textbf{22.1} & \textbf{60.7} & 89.4 & 66.3 & 61.6 & 83.8 & \textbf{69.2} & 64.7 \\
    CoCoOp & 20.1 & 59.8 & \textbf{90.4} & \textbf{67.9} & \textbf{63.8} &  \textbf{87.6} & 69.1 & \textbf{65.5} \\
    \midrule
    \textbf{ResNet-101} & \\
    CLIP & 17.5 & 63.2 & 89.5 & 62.4 & 62.2 & 84.2 & 61.3 & 62.9 \\
    CoOp & \textbf{24.6} & \textbf{68.2} & 92.0 & 68.3 & 65.4 & 88.2 & \textbf{72.7} & \textbf{68.5} \\
    CoCoOp & 22.5 & 65.2 & \textbf{93.3} & \textbf{69.9} & \textbf{67.5} & \textbf{88.6} & 71.5 & 68.4 \\
    \midrule
    \textbf{ViT-B/32} & \\
    CLIP & 18.2 & 60.1 & 91.6 & 61.3 & 61.8 & 85.5 & 61.3 & 62.8 \\
    CoOp & \textbf{24.0} & \textbf{63.0} & 93.6 & 67.3 & 65.7 & \textbf{88.5} & \textbf{74.5} & \textbf{68.1} \\
    CoCoOp & 19.5 & 60.4 & \textbf{93.8} & \textbf{69.8} & \textbf{67.3} & \textbf{88.5} & 72.7 & 67.4 \\
    \midrule
    \textbf{ViT-B/16} & \\
    CLIP & 24.4 & 64.9 & 92.6 & 67.5 & 67.9 & 87.4 & 66.1 & 67.2 \\
    CoOp & \textbf{32.4} & \textbf{72.4} & 94.7 & 73.2 & 72.1 & 90.1 & \textbf{78.2} & \textbf{73.3} \\
    CoCoOp & 30.4 & 68.7 & \textbf{94.8} & \textbf{73.5} & \textbf{73.6} & \textbf{91.6} & 76.3 & 72.7 \\
    \bottomrule
    \end{tabular}
\end{table*}

\section{Results on DOSCO-2k} \label{appx:dosco_2k}

\paragraph{DOSCO-2k}
The DOSCO (DOmain Shift in COntext) benchmark~\citep{zhou2022device} contains 7 image recognition datasets, which cover a wide range of classification problems, such as generic object recognition, fine-grained recognition on aircraft models, and action recognition. Unlike existing domain generalization datasets where the domain labels are manually defined and often limited to image style variations, DOSCO-2k focuses on broader contextual domain shift, which is automatically detected by a neural network pre-trained on the Places dataset~\citep{zhou2017places}. Following~\citet{zhou2022device}, we use the 2k version where the training and validation splits in each dataset have 2,000 images in total (1,600 for training and 400 for validation).

\paragraph{Results}
We study three methods' domain generalization performance on DOSCO-2k: CLIP, CoOp and CoCoOp~\citep{zhou2022conditional}. All models are trained on the training set and the checkpoints with the best validation performance are used for final test in unseen domains. Table~\ref{tab:dosco_2k} shows the results of four different architectures. It is clear that the two learnable methods outperform the zero-shot method with a large margin, despite having only a small number of parameters to tune. CoCoOp beats CoOp on 4 out of 7 datasets but CoOp's average performance is higher. In summary, the results suggest that efficient adaptation methods like CoOp and CoCoOp have great potential in tackling transfer learning problems.


%
%

\bibliographystyle{spbasic}      
\bibliography{ref}   

\end{document}